\pdfoutput=1

\PassOptionsToPackage{table, prologue}{xcolor}
\PassOptionsToPackage{backref=page}{hyperref}

\documentclass[11pt]{article}

\usepackage[]{EMNLP2022}

\usepackage{times}
\usepackage{latexsym}
\usepackage{graphicx}
\usepackage{booktabs}
\usepackage{pgfplots}
\pgfplotsset{compat=newest}
\usepgfplotslibrary{colorbrewer}
\usepackage[backgroundcolor=cyan,linecolor=black]{todonotes}

\usepackage{xspace}

\newcommand{\sref}[1]{\S\ref{#1}}

\newcommand{\cameraready}[1]{}

\usepackage[T1]{fontenc}

\usepackage[utf8]{inputenc}

\usepackage{microtype}

\usepackage{inconsolata}
\newcommand{\mysubsection}[1]{\vspace{0.3em}\noindent\textbf{#1}}

\title{A Critical Reflection and Forward Perspective on\\Empathy and Natural Language Processing}

\author{Allison Lahnala$^{1}$, Charles Welch$^{1,3}$, David Jurgens$^{2}$, Lucie Flek$^{1,3}$ \\
    $^{1}$Conversational AI and Social Analytics (CAISA) Lab \\ 
    Department of Mathematics and Computer Science, University of Marburg  \\
    $^{2}$ School of Information, University of Michigan \\
    $^{3}$ The Hessian Center for Artificial Intelligence (Hessian.AI) \\
    \texttt{\{allison.lahnala,welchc,lucie.flek\}@uni-marburg.de}, \texttt{jurgens@umich.edu} 
}

\begin{document}
\maketitle
\begin{abstract}

We review the state of research on empathy in natural language processing and identify the following issues: (1) empathy definitions are absent or abstract, which (2) leads to low construct validity and reproducibility. Moreover, (3) emotional empathy is overemphasized, skewing our focus to a narrow subset of simplified tasks. We believe these issues hinder research progress and argue that current directions will benefit from a clear conceptualization that includes operationalizing cognitive empathy components. Our main objectives are to provide insight and guidance on empathy conceptualization for NLP research objectives and to encourage researchers to pursue the overlooked opportunities in this area, highly relevant, e.g., for clinical and educational sectors.

\end{abstract}

\section{Introduction}

Interest in empathetic language continues to grow in natural language processing research. 
Empathy recognition and empathetic response generation tasks have become well-established research directions, especially since the introduction of now fairly mainstream benchmark datasets for each task, \textsc{Empathic Reactions} \cite{buechel-etal-2018-modeling} and \textsc{Empathetic Dialogues} \cite{rashkin-etal-2019-towards}, and an empathy detection shared task established on the former  \cite{tafreshi-etal-2021-wassa,barriere-etal-2022-wassa}.

These empathy-focused tasks are highly motivated by myriad benefits, including (i) improved experiences with conversational agents~\cite{shuster-etal-2020-dialogue,roller-etal-2021-recipes} and satisfaction with customer care dialogue agents \cite{firdaus-etal-2020-incorporating,sanguinetti-etal-2020-annotating}, (ii) computational social science analyses, e.g., supportive interactions in online forums~\cite{khanpour-etal-2017-identifying,zhou-etal-2020-design,sharma-etal-2020-computational}, and (iii) tools to assist in training and evaluating health practitioners~\cite{demasi-etal-2019-towards,perez-rosas-etal-2017-understanding}. 
With this increased interest in empathy-focused NLP, however, has come little clarity on what empathy is and how it is being operationalized. Papers vary substantially in how they define, annotate, and evaluate empathy, leading to a fractured landscape that hinders progress.

Most NLP work has loosely defined empathy as the \textit{ability to understand another person's feelings and respond appropriately}. As a result, these works have focused primarily on detecting sentiment and emotions in text as proxies for \textit{understanding feelings} and consider empathetic responses to be those that demonstrate success in emotion recognition and express so with a tone consistent with the valence of the target's sentiment. 
Under this view, systems that primarily recognize or perform emotionally coherent interactions supposedly fulfill the goal of having an empathetic system. However, established findings and theories about human empathy show that this objective is short-sighted---or even misdirected---and misses a critical system  known as \textit{cognitive empathy}.

We argue that NLP research has omitted key aspects of empathy through its narrow focus on emotion and, as a result, led us to neglect the cognitive components. Our paper offers three key contributions. First, to show this gap in research on empathy, we  provide a theoretical grounding for empathy from psychology (\sref{sec:theory}). We then survey empathy literature in NLP focusing on ACL$^{*}$ venues (\sref{sec:lit_review}) and highlight three central problems: (1) computational work has overlooked much of empathy through vague definitions and a focus on emotion (\sref{sec:definition-problem}), (2) our current underspecification of empathy leads to issues in construct and data validity (\sref{sec:measurement-problem}), and (3) the narrow empathy tasks we choose as a community limit our progress (\sref{sec:task-problem}). Finally, we propose a way forward (\sref{sec:refocus}) through a clear conceptualization of empathy that operationally considers processes of cognitive empathy and highlight overlooked research opportunities (\sref{sec:going-forward}).

\section{What \textit{is} Empathy? A Theoretical Guide}
\label{sec:theory}

Empathy is a multi-dimensional construct that contains both emotional and cognitive aspects which relate to how an observer reacts to a target~\citep{davis1980-multidim-empathy,davis1983measuring}. In practice, empathy is diversely defined in terms of these social, emotional, cognitive, and even neurological dimensions \cite{cuff2016empathy}. Indeed, while folk conceptions of empathy refer to a single construct, multiple studies have pointed to distinct neurological systems for emotional and cognitive empathy \citep{decety2006social,shamay2011neural}.

In describing empathy in communication, we adopt the standard terminology of (i) a \textit{target} as someone experiencing an emotion or situation and (ii) an \textit{observer} as another person at the disposition for an empathic experience through perceiving the target's emotions and situation.

In psychology, the most discussed aspects of empathy are its affective and cognitive components \cite{cuff2016empathy}. The affective components, referred to as \textit{emotional empathy}, relate to the observer's emotional reaction to the target \cite{shamay2011neural}. The cognitive components, referred to as \textit{cognitive empathy}, relate to active processes used by the observer to infer the mental state of the target~\cite{blair2005responding}. Emotional empathy represents \textit{automatic} (bottom-up) processes whereas cognitive empathy represents \textit{controlled} (top-down) processes~\cite{lamm2007neural}, and they interact with each other.

\subsection{Emotional Empathy}\label{sec:emotional_empathy}

Emotional empathy, or the observer's capacity to experience affective reactions upon perceiving the target, can involve processes such as emotion recognition, contagion, and pain sharing \cite{shamay2011neural}.\footnote{We further discuss contagion and mimicry in Appendix~\ref{appendix:contagion_mimicry}.} When these affective processes caused by perceiving the target's emotional state interact with certain contextual factors, the observer's experience is a form of \textit{emotional empathy}. Such forms include the concepts of sympathy, compassion, and tenderness. In NLP, we often use these terms to define empathy without any distinction between them. Accordingly, our review finds we do not investigate them as specific empathy-related phenomena or characterizations of empathetic responses. As we describe in this section, however, the degree to which specific contextual factors interact with the internal affective processes renders these concepts distinct. We propose differentiable characteristics of sympathy, compassion, and tenderness based on psychology literature. Furthermore, these characteristics are operationalizable for more precise NLP research on emotional empathy.

\paragraph{Distinguishing sympathy, compassion, and tenderness.} While each of these concepts relates to having feelings \textit{for} the target, each one is distinct regarding the perceived immediacy of the target's needs, vulnerability, and a desire to help. 
Sympathy relates to the present situation of the target and involves sorrow and concern for the target arising from perceived suffering, whereas tenderness relates to their long-term needs, and involves warm and fuzzy feelings from perceiving the target as vulnerable, delicate, or defenseless. Compassion is a higher construct that involves feelings based on the target's perceived needs and motivated desires to protect the vulnerable and provide care to those who suffer~\cite{goetz2010compassion}.

Feeling \textit{for} the target means emotions can be different between the target and observer. Emotional congruence is used to describe when an emotion is felt by both target and observer, though some define congruence as a response to a situation that is similar but may result in a different emotion. In this case, the emotion is congruent if it is appropriate for the situation. This idea plays a significant role in how the NLP community views empathetic responses, though we consider it unnecessary for empathy in general.

We can perceive another person as vulnerable or suffering and experience tenderness, sympathy, and compassion for them without thinking hard about it. In other words, cognitive processes are not strongly required for these experiences, distinguishing them as \textit{emotional empathy}. However, empathic cognitive processes can help render these experiences by elevating awareness of those contextual factors (i.e., realizing the target's needs and vulnerability) through active deliberation of the target's situation. In the next section, we describe such processes of \textit{cognitive empathy}.

\subsection{Cognitive Empathy}\label{sec:cognitive_empathy}

Cognitive empathy centers, in part, on perspective-taking, the process of the observer conceptualizing the target's point of view. This active process is the primary way of achieving cognitive empathy, though other scenarios such as imagined memories or fictional scenarios may also be processes that help to achieve it~\cite{eisenberg2014altruistic,stinson1992empathic}.

Psychologists have proposed a variety of frameworks for what actions or processes constitute perspective-taking. For example, in the \textit{appraisal theory of empathy} \citep{wondra2015appraisal}, empathy is considered with respect to different aspects or dimensions of perspective that an observer might have for the target's situation. Six different types of appraisals are proposed: pleasantness, anticipated effort to deal with the situation, anticipated control of the situation, self-other agency for responsibility of the situation, attentional activity (degree of surprise), and certainty of the situation or outcome.
With this view, how ``empathetic'' the observer is depends on the number of appraisals  and the degree to which their responses or actions mirror the true feelings of the target. 

To be able to empathize successfully, the observer needs to be able to accurately infer the content of the target's thoughts and feelings \cite{ickes2011everyday}. The degree of accuracy in the observer's inferences is known as \textit{empathic accuracy}. High empathic accuracy is particularly essential in clinical domains, such as Motivational Interviewing \citep{miller2012motivational}, where the therapist (observer) formulates \textit{reflections} based on interpreting what the patient (target) says (see examples in Table~\ref{tab:complex_reflections}). 
 
When considering the factors that affect empathic accuracy, the observer is not the only possible source of error. The target must also express and convey the situation accurately. As a result, it is unlikely that both the observer and target will ever perceive situations exactly the same~\cite{stotland19780}. This limitation has implications for empathy annotations, with the inter-annotator agreement being subject to the empathic accuracy and subjective interpretations of the annotators. The \textit{unstructured dyadic interaction} and \textit{standard stimulus} paradigms are two study approaches developed in interpersonal perception research for measuring empathic accuracy that involve comparing observer inferences with the target self-reports of their actual thoughts and feelings \cite{ickes2001measuring}; such approaches to study empathic accuracy have (to our knowledge) never been explored in NLP research.

\begin{table}[]
    \centering
    \small
    \rowcolors{1}{gray!15}{white}
    \begin{tabular}{rp{5.4cm}}
    \toprule
        \textbf{Client} & Well, I know I need to stay active. ``Use it or lose it,'' they say. I want to get my strength back, and they say regular exercise is good for your brain, too. \\
        \textbf{Interviewer} & So that's a puzzle for you -- how to be active enough to get your strength back and be healthy, but not so much that would put you in danger of another heart attack. \\
        \textbf{Client} & I think I'm probably being too careful. My last test results were good. It just scares me when I feel pain like that. \\
        \textbf{Interviewer} & It reminds you of your heart attack.\\
        \textbf{Client} & That doesn't make much sense, does it-- staying away from activity so I won't have another one?\\
        \textbf{Interviewer} & Like staying away from people so you won't be lonely.\\
        \textbf{Client} & Right. I guess I just need to do it, figure out how to gradually do more so I can stick around for a while.\\
    \bottomrule
    \end{tabular}
    \caption{A motivational interviewing interaction demonstrating complex reflections. Example from \cite{miller2012motivational}.}
    \label{tab:complex_reflections}
\end{table}

\begin{table*}
\centering
\rowcolors{2}{gray!15}{white}
\begin{tabular}{lp{.81\linewidth}r}
\toprule
 & \textbf{Definition Themes} &  \textbf{Count} \\
\midrule
D0 & Studies do not define or describe empathy. &      8 \\
D1 & Studies do not describe empathy but an abstract conceptualization could be inferred from the task or system description. &     10 \\
D2 & Studies provide empathy descriptions that are vague or ambiguous with other concepts. &     10 \\
D3 & Studies describe relationships between certain behaviors and empathy. &      3 \\
D4 & Studies provide succinct yet explicitly theory-grounded empathy descriptions and adhere to the theoretical foundation. &      9 \\
D5 & Studies provide thorough, theory-based descriptions of empathy. &      8 \\
\bottomrule
\end{tabular}
\caption{The number of papers identified for each Definition Theme in our review.}
\label{tab:def_themes}
\end{table*}

\section{Literature Review: Empathy in NLP}\label{sec:lit_review}

Empathy research in NLP primarily focuses on empathy recognition and empathetic response generation. There is also work that analyzes empathy-related behaviors, such as supportive intents and counselor strategies. Across this literature, we have found highly varied and often underspecified usage of the term \textit{empathy}.

We summarize our findings from reviewing a collection of computational linguistics and natural language processing papers. Papers were identified by searching for the term ``empath" in the ACL anthology. Then we narrowed the resulting sample from 90 to 69 papers whose investigation involves empathy, as opposed to mentioning ``empathy'', ``empathetic'', or ``empathic'' rhetorically. We also included a selection of relevant works outside the ACL anthology from venues such as AAAI. 

We focused on papers presenting systems for empathy prediction or empathetic response generation, evaluations of empathy dialogue models, empathy annotation schemes and datasets, and corpus analyses of empathetic language. We labeled a total of 48 papers (excluding 14 WASSA\footnote{Workshop on Computational Approaches to Subjectivity, Sentiment \& Social Media Analysis} shared task papers) based on their descriptions of empathy and empathy evaluation approaches. 

\subsection{Findings}

We identified six predominant themes of \textit{empathy descriptions} (D0-5), and seven \textit{empathy evaluation approaches} (E0-6). Tables \ref{tab:def_themes} and \ref{tab:empathy_eval_labels} show the categories we defined for grouping the description and evaluation themes, together with the number of papers we identified characterized by these themes. 

The definition themes are as follows (details and examples are provided in Appendix~\ref{appendix:definition_themes}):

\mysubsection{D0: Studies do not define or describe empathy.} Despite the significance of empathy in such studies, a conceptualization is not given nor can it be reasonably inferred. In some cases, empathy is associated with \textit{politeness} and \textit{courtesy}.

\mysubsection{D1: Studies do not describe empathy but an abstract conceptualization could be inferred from the task or system description.} This is especially the case among empathetic response generation research. In task descriptions, the tendency is to describe empathetic response generation as \textit{the task of understanding the user's emotions and responding appropriately}. System descriptions can be informative about the grounding concept when the design reflects specific dimensions of empathy. For instance, an empathetic response generation system that mainly relies on an emotion recognition module for the purpose of response conditioning suggests the significance of emotion understanding in the work’s conceptualization.

\mysubsection{D2: Studies provide empathy descriptions that are vague or ambiguous with other concepts}. Some studies use \textit{sympathy} and \textit{empathy} interchangeably, 
and others nearly exchange the term \textit{empathy} with \textit{emotion recognition} by strongly associating them without providing disambiguation.
These often regard an appropriate response as one that mimics or mirrors the target's response. This characterization usually accompanies a system that conditions a response on emotions or sentiments predicted by a dedicated module. 
In other cases, papers reference the conceptualization linked to the dataset they used, as is frequent among WASSA shared task papers (listed in Appendix~\sref{appendix:evaluation_categories}) using the \textsc{Empathic Reactions} dataset \cite{buechel-etal-2018-modeling}. 

\mysubsection{D3: Studies describe relationships between certain behaviors and empathy.} These works, mainly concerned with narrative and conversation analyses, provide thorough descriptions of other concepts and their relations to empathy that are consistent with multiple aspects of empathy described in \sref{sec:theory}.

\mysubsection{D4: Studies provide succinct yet explicitly theory-grounded empathy descriptions and adhere to the theoretical foundation.} This research often involves a counseling/therapy conversations dataset with labels from a scheme developed by experts in those domains.

\mysubsection{D5: Studies provide thorough, theory-based descriptions of empathy.} This is often the case in works that develop a novel multi-dimensional empathy framework for analysis and annotations. As in D4, they also tend to involve experts in behavioral, social, and health domains.

\begin{table*}
\centering
\rowcolors{2}{gray!15}{white}
\begin{tabular}{lp{.55\linewidth}r}
\toprule
 & \textbf{Evaluation Themes} &  \textbf{Count} \\
\midrule
E6 & Multi-item: cognitive \& emotional &      8 \\
E5 & Single label/rating: cognitive \& emotional empathy &      3 \\
E4 & Single label/rating: only emotional empathy &     19 \\
E3 & Single label/rating: no specification &      8 \\
E2 & Heuristic empathy labels/ratings &      4 \\
E1 & Target-observer role labeling &      2 \\
E0 & No manual evaluation or only automatic &      4 \\
\bottomrule
\end{tabular}
\caption{Themes of evaluations and annotations of empathetic language in NLP literature with counts of papers described the theme. Detailed descriptions of the themes are provided in Appendix~\ref{appendix:evaluation_categories}.}
\label{tab:empathy_eval_labels}
\end{table*}

\section{The Definition Problem}
\label{sec:definition-problem}

The lack of delineation between empathy and related concepts in psychology ultimately leaves us wondering what we are studying. When empathy is not explicitly conceptualized (D0-1), the training data is often implicitly focused on components of emotional empathy and responses based on emotion-matching strategies. However, since the data properties are often under-specified in the NLP papers as well, we only have indirect proxies to infer such emotion-centric conceptualizations. In an ideal case, the training data description would reveal more about the relevant features of emotion matching/mirroring, e.g., separating contextual factors that would define a particular emotional empathy behavior such as sympathy, compassion, or tenderness. 

While empathy is indeed complex, we demonstrated in Section~\ref{sec:emotional_empathy} a way to disambiguate these empathetic behaviors by considering the variables of how the observer perceives the target's vulnerability and needs. These are just some possible manifestations of emotional empathy that NLP researchers could investigate using more detailed study designs to control such variables. To alleviate the problem of abstractness and potential inconsistencies, we recommend that researchers disambiguate their objectives from simply ``empathy'' by considering such concepts as sub-areas of the empathy research direction. 

\begin{figure}[t]
    \centering

    \begin{tikzpicture}
      \begin{axis}[enlargelimits=false,width=6.7cm,colorbar,colormap/Greys,
    xtick = {0,1,2,3,4,5,6},
    xticklabels = {E0,E1,E2,E3,E4,E5,E6},
    ytick = {0,1,2,3,4,5},
    yticklabels = {D5,D4,D3,D2,D1,D0},
    ylabel = {Definition Themes},
    xlabel = {Evaluation Themes},
    every node near coord/.append style={yshift=-0.25cm} 
      ]
        \addplot [matrix plot,point meta=explicit, ]
          coordinates {
(0,0) [0] (1,0) [1] (2,0) [0] (3,0) [1] (4,0) [3] (5,0) [1] (6,0) [2] 

(0,1) [0] (1,1) [1] (2,1) [1] (3,1) [1] (4,1) [3] (5,1) [0] (6,1) [3] 

(0,2) [0] (1,2) [0] (2,2) [0] (3,2) [0] (4,2) [1] (5,2) [0] (6,2) [2] 

(0,3) [1] (1,3) [0] (2,3) [1] (3,3) [2] (4,3) [4] (5,3) [1] (6,3) [1] 

(0,4) [0] (1,4) [0] (2,4) [1] (3,4) [3] (4,4) [5] (5,4) [1] (6,4) [0] 

(0,5) [3] (1,5) [0] (2,5) [1] (3,5) [1] (4,5) [3] (5,5) [0] (6,5) [0] 
 
          };
        \node[text=black] at (axis cs: 0,0) {0};
        \node[text=black] at (axis cs: 1,0) {1};
        \node[text=black] at (axis cs: 2,0) {0};
        \node[text=black] at (axis cs: 3,0) {1};
        \node[text=black] at (axis cs: 4,0) {3};
        \node[text=black] at (axis cs: 5,0) {1};
        \node[text=black] at (axis cs: 6,0) {2};
        \node[text=black] at (axis cs: 0,1) {0};
        \node[text=black] at (axis cs: 1,1) {1};
        \node[text=black] at (axis cs: 2,1) {1};
        \node[text=black] at (axis cs: 3,1) {1};
        \node[text=black] at (axis cs: 4,1) {3};
        \node[text=black] at (axis cs: 5,1) {0};
        \node[text=black] at (axis cs: 6,1) {3};
        \node[text=black] at (axis cs: 0,2) {0};
        \node[text=black] at (axis cs: 1,2) {0};
        \node[text=black] at (axis cs: 2,2) {0};
        \node[text=black] at (axis cs: 3,2) {0};
        \node[text=black] at (axis cs: 4,2) {1};
        \node[text=black] at (axis cs: 5,2) {0};
        \node[text=black] at (axis cs: 6,2) {2};
        \node[text=black] at (axis cs: 0,3) {1};
        \node[text=black] at (axis cs: 1,3) {0};
        \node[text=black] at (axis cs: 2,3) {1};
        \node[text=black] at (axis cs: 3,3) {2};
        \node[text=white] at (axis cs: 4,3) {4};
        \node[text=black] at (axis cs: 5,3) {1};
        \node[text=black] at (axis cs: 6,3) {1};
        \node[text=black] at (axis cs: 0,4) {0};
        \node[text=black] at (axis cs: 1,4) {0};
        \node[text=black] at (axis cs: 2,4) {1};
        \node[text=black] at (axis cs: 3,4) {3};
        \node[text=white] at (axis cs: 4,4) {5};
        \node[text=black] at (axis cs: 5,4) {1};
        \node[text=black] at (axis cs: 6,4) {0};
        \node[text=black] at (axis cs: 0,5) {3};
        \node[text=black] at (axis cs: 1,5) {0};
        \node[text=black] at (axis cs: 2,5) {1};
        \node[text=black] at (axis cs: 3,5) {1};
        \node[text=black] at (axis cs: 4,5) {3};
        \node[text=black] at (axis cs: 5,5) {0};
        \node[text=black] at (axis cs: 6,5) {0};
      \end{axis}
    \end{tikzpicture}
    \caption{Heatmap of definition and evaluation themes.}
    \label{fig:heatmap}
\end{figure}
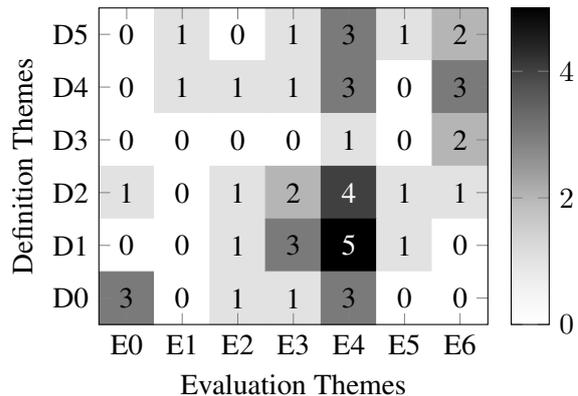

Nearly thirty studies in our review provide vague or no empathy descriptions (D0-2). Without the ability to focus on individual aspects or an understanding of the nature of empathetic interactions and thus what constitutes ``appropriate'' responses, the issue of what we are measuring arises. Figure \ref{fig:heatmap} displays the interaction between empathy definitions and evaluation designs. There is a clear correspondence between under-specified empathy descriptions and under-specified evaluations. 
Even when a study provides a multi-dimensional definition of empathy (D4-5), it often is not reflected in the evaluation design. Rather the evaluations default intuitively to the emotional components (E4). 

The undefined nature of empathy ultimately manifests in poor operationalization and may contribute to inconsistency in how empathy is measured from observations, resulting in poor construct validity \cite{coll2017we}. With unstable validity, these works may not be deemed reliable for clinical applications, such as counselor training, which has been noted as a critical application of artificial intelligence to the field of psychotherapy~\cite{imel2017technology}. Aside from such applications, poor validity and lack of definitions lead to reproducibility issues \cite{cuff2016empathy}. Lacking a shared understanding and robust conceptualization of empathy makes it difficult to interpret findings and compare studies.

\section{The Measurement Problem}
\label{sec:measurement-problem}

Given the overwhelmingly abstract portrayal of empathy, the effectiveness and validity of our approaches for measuring, annotating, and evaluating it are questionable. Simply put, we do not know if we are investigating the same thing or doing so consistently. Following, we outline where NLP needs to improve in its measurements to move forward.

\paragraph{Construct validity implications for resource construction and evaluation.}
\citet{yalcin2019evaluating} reviews several scales developed in psychology and suggests potential adaptations for evaluating empathy in NLP.
However, issues with measurement, construct, and predictive validity also persist in psychology \cite{ickes2001measuring} and reflect similarly in the limitations of NLP research, e.g., ambiguous evaluation~\cite{coll2017we}.

Established psychological scales can help measure empathy, such as the often used Empathic Concern and Personal Distress Scale~\cite{batson1987distress,buechel-etal-2018-modeling}.
One could, for instance, ask participants to read a passage written by an individual and ask the reader about their emotions. However, various non-empathetic factors can affect how someone feels after reading, not resulting from perspective-taking or understanding the target's emotional state. Psychological studies that use these scales devise experimental controls to promote an other-focus state~\cite{fabi2019empathic,batson1997perspective}. One method is to instruct participants to ``imagine how the person ... feels about what has happened and how it has affected his or her life''~\cite{toi1982more}. Another is to control how similar the observer perceives themselves to be to the target~\cite{batson1981empathic}.
Self-report measures such as the Davis IRI scale \cite{davis1983measuring} can measure empathic capabilities, but empathy still varies across situations~\cite{litvak-etal-2016-social,cuff2016empathy}.

Empathy involves an accurate understanding of another's mental state. As such, one can compare descriptions given by an observer and a target~\cite{ickes2001measuring}. These first-person assessments are more accurate than annotations by a third party attempting to judge mental states from language, behaviors, or situational factors.

\mysubsection{Empathy and domain shift: What exactly are we trying to transfer?}
The scarcity and diversity of existing datasets motivate investigating knowledge transfer between them~\cite{wu-etal-2021-towards,lahnala-etal-2022-caisa}.
Domains vary across datasets, but datasets also vary in how they capture empathy. These differences make larger-scale efforts that combine datasets more difficult. Some studies leverage heuristics to curate empathy data (see Appendix \ref{appendix:evaluation_categories}), such as bootstrapping with pre-trained models, selecting interactions from particular contexts (e.g., particular subreddits), and crowd-sourcing conversations grounded on particular emotions. While heuristic approaches can help mitigate curation costs \cite{hosseini2021takes, welivita-etal-2021-large,wu-etal-2021-towards}, the effectiveness is subject to their validity. 

In a study motivated to minimize annotation effort through domain transfer, for instance, \citet{wu-etal-2021-towards}'s experimental results demonstrated the heuristically curated datasets, \textsc{Empathetic Dialogues} and \textsc{PEC}~\cite{zhong-etal-2020-towards}, were \textit{insufficient} for predicting empathy in the Motivational Interviewing domain. They suggest that more fine-grained empathy annotation labels would help smooth the domain gaps by distinguishing the empathy aspects expected to be present. Such efforts are needed for the community to build off of each other's work. Meanwhile, further investigations of knowledge transfer techniques between existing resources could have positive contributions.\cameraready{\footnote{We share preliminary investigations on English empathy-related datasets in Appendix \sref{appendix:knowledge_transfer} based on \citet{poth-etal-2021-pre}'s techniques for intermediate task selection in Appendix. Cross-lingual transfer with non-English datasets is another direction to investigate \cite{pfeiffer-etal-2020-mad}.}}

\section{The Narrow Task Problem}\label{sec:more_than_emotions}
\label{sec:task-problem}

While empathy is widely recognized as a construct NLP should be modeling, we argue that the field has focused on a narrow set of tasks that have held back progress. By bridging current work with a perspective of cognitive empathy, we motivate a series of new and reimagined tasks that would advance the field's ability to model empathy.

\mysubsection{Empathy is more than emotions.}
The predominant empathy conceptualization requires the observer to understand the target's emotions and respond appropriately. We recommended that researchers disambiguate sympathy, compassion, and tenderness and view these as subareas of empathy research. However, we argue that this research would still suffer by ignoring cognitive empathy and its interaction with emotion understanding and empathetic responses. Some perspectives from psychology emphasize that the distinction between emotional and cognitive empathy is less important than the interaction between the two components \cite{cuff2016empathy}--and we can draw inspiration from that. Cognitive empathy processes are necessary considerations that can more effectively achieve goals for the emotional empathy subareas. 

\mysubsection{Cognitive empathy processes can improve methods for \textit{understanding emotions}.} 
In NLP, we often consider emotions first, assuming they can be inferred directly from what the target expresses. However, the target could minimize expressions, making the emotions harder to perceive (refer to the discussion on \textit{empathic accuracy} \sref{sec:cognitive_empathy}). Furthermore, most papers only operate on text (e.g., transcripts), which misses opportunities for empathetic cues in other modalities. 

Cognitive processes help with better emotional understanding yielding higher empathic accuracy of inferences about the target's affective state.
\citet{cuff2016empathy} argue that the observer can \textit{infer} emotionality through perspective-taking, imagination, and retrieval of relevant memories. In this way, a cognitive approach first leads to an understanding and more accurate perception of emotions.

We argue that cognitive empathy can benefit from approaches such as common sense reasoning~\cite{sabour2021cem}, external knowledge~\cite{li2020towards}, and abductive NLI~\cite{bhagavatula2020abductive}.
For example, \citet{shen-etal-2021-constructing-emotional} integrated external knowledge to improve counselor reflection generation. \citet{tu-etal-2022-misc} presented an approach to understanding the target's emotional state using commonsense knowledge that improves over more emotion-focused models. Their approach was inspired by \citet{ma-2020-survey}'s survey of empathetic dialogue systems, which argued that future work should go beyond emotional empathy by pursuing personalization and knowledge (both contextual and general external knowledge. Integrating these types of knowledge supports \textit{reasoning about emotion cause}, as opposed to only recognizing emotions, which itself is insufficient~\cite{gao-etal-2021-improving-empathetic,kim-etal-2021-perspective}. Models that incorporate reasoning about the cause of emotion were shown to outperform emotion recognition and mirroring counterparts, notably including those that included external commonsense and emotional lexical knowledge, by human evaluated empathy scores, i.e. \cite{lin-etal-2019-moel,li-etal-2020-empdg,majumder-etal-2020-mime}.

Studies may even be able to specify and investigate what type of information is necessary in order to understand the user's emotions and model empathically accurate responses. This can be done with a controlled selection of samples from knowledge bases. For instance, \cite{shen-etal-2020-counseling,shen-etal-2022-knowledge}'s work uses particular aspects of common sense and domain-specific knowledge. Researchers must become more intimate with data and understand the nature of dialogue to design empathetic systems.

\mysubsection{Cognitive empathic processes can improve methods to select \textit{appropriate response} strategies.} Approaches inspired by cognitive processes could thus result in not only valuable representations of emotion but also enhanced methods for response strategies. Though we critiqued the generally-lacking specificity about response appropriateness earlier, the most salient idea we grasp is that observers should mirror the target's emotion or express a similar sentiment valence.
Contrary to this idea, \citet{xie-pu-2021-empathetic} find that empathetic listeners often respond to negative emotions such as sadness and anger with questions rather than expressing similar or opposite emotions. In addition, specific empathetic question intents of observers may play more significant roles in regulating the target's emotions~\cite{svikhnushina-etal-2022-taxonomy}. Approaches designed both on emotional and cognitive schemes also had effective results in assisting students write peer-reviews that are perceived more empathetic~\cite{wambsganss-etal-2021-supporting,wambsganss-etal-2022-adaptive}. 

\section{Refocusing our efforts}
\label{sec:refocus}

\mysubsection{We overlook research opportunities that better align with described motivations, and these problems necessitate cognitive empathy.} Most empathy research in NLP focuses on emotional chatbots or social media analysis. This focus 
probably drives the overemphasis on emotional empathy. Neither of the domains, we argue, is helpful for the purposes where empathetic NLP systems would be needed the most; for clinical or educational praxis or other rather formal contexts. Such scarce applications include research on Motivational Interviewing  \cite{perez2016building,perez-rosas-etal-2017-understanding,perez-rosas-etal-2018-analyzing,perez-rosas-etal-2019-makes,wu-etal-2021-towards,shen-etal-2022-knowledge}, assistive writing systems for peer-to-peer mental health support \cite{sharma2021towards,sharma-etal-2022-collab}, and other counseling conversations \cite{althoff-etal-2016-large,zhang-danescu-niculescu-mizil-2020-balancing,zhang2020quantifying,sun-etal-2021-psyqa}. Work in these areas make the need for cognitive empathy even more apparent.

\cameraready{Current methods assume that the observer's ability to infer the target's emotion or cognitive state is fixed. However, empathy is partly a learned skill; prior work has shown a distinction between trait and state empathy \citep{kunyk2001clarification}---trait relates to the empathic abilities and state relates to contextual factors. Trait empathy may be honed through training, suggesting that the cognitive processes involved in perspective-taking can be learned and improved \cite{gerdes2011teaching}. }

\cameraready{Must empathy always be linked to emotion? If emotion is left aside, the remaining process aims to understand a target's thoughts and the sequence of events that led to the present. Computational models just for this component could be beneficial in many scenarios that do not necessarily require an emotional component in the observer (e.g. therapists, teachers, medical professionals). In these primarily-cognitive settings,  we may instead use the term \textit{empathic understanding}~\cite{rogers1967therapeutic}. }

Empathic capacities vary across individuals, as some experience more intense affective responses to certain target perceptions than others~\cite{eisenberg2014altruistic}. Without techniques to manage affective responses, people can experience significant distress. Cognitive empathy skills strengthen observers' ability to regulate or manage their affective responses. Thus, these skills are critical for individuals in caregiving roles (e.g., counselors and doctors) who interpersonally engage with others who are vulnerable, suffering, or in crisis daily. At the same time, the ability to empathize with the target is essential for their roles in providing effective treatment, making the role of cognitive empathy significant both for self-care and care for others.

NLP research is highly needed for \textit{virtual standardized patient} (VSP) systems. VSPs provide an effective way for learners to develop cognitive empathy and clinical skills~\cite{Lok2019-VSP-teach-empathy}. The need for VSPs has grown from the necessity for methods that allow learners to practice rare and realistic crisis scenarios and concern for the safety and expenses of standardized patients. Needs in this area include but are not limited to conversational models for practice in intercultural communication, end-of-life discussions, and breaking bad news. Development of such systems naturally requires the skills of NLP researchers, and these research efforts would benefit the field. 

\cameraready{\cite{Lok2019-VSP-teach-empathy} Virtual standardized patient use has grown due to concerns of safety and expense related to standardized patients, emphasis on self-learning, and expansion of the distributed education model, calling for standardization of instruction for learners enrolled in the same educational program on distant campuses.
In medicine, virtual standardized patients are described as multimedia interactive scenarios that allow safe practice and repetition and immediate feedback, help develop clinical skills, and can simulate rare but critical scenarios [7, 9, 10, 13, 17]. While classroom-based and online courses can effectively deliver course content, they cannot provide personalized learning for each trainee. Objective, individualized feedback given to each trainee is crucial for learning [17–19]}

So far, the task of modeling and simulating the empathy target has received little attention in NLP, leaving a significant research gap that our community is positioned to fill, with exciting challenges that align with the motivations of research on empathy and that can have a broad impact within and beyond our field. One work, for instance, focused on simulating individuals in crisis for the use case of counselor training~\cite{demasi-etal-2019-towards}. Similarly, modeling and simulating the target could support broader educational objectives, such as training students' emotionally and cognitively empathic feedback skills \cite{wambsganss-etal-2021-supporting}. This direction of simulating a mental health support seeker aligns with the goals of related work for developing tools to assist with evaluating and training counselors~\cite{perez-rosas-etal-2018-analyzing,perez-rosas-etal-2019-makes,imel2017technology,zhang-danescu-niculescu-mizil-2020-balancing}. The ethics section further discusses our perspective on this approach compared to developing empathetic response generation models for support-related motivations.

\section{A Forward Perspective}
\label{sec:going-forward}

Here we summarize the main obstacles we identified and our recommendations for moving forward. 

\mysubsection{The Definition Problem: Define research goals and design methodologies based on empathy's more concrete and measurable aspects.} The abstractness of empathy has negative implications for the construct validity of measurement approaches and, thereby, the scientific effectiveness and comparability. 

\mysubsection{The Measurement Problem: Draw inspiration from measurements established in psychology, and in addition, learn from investigations and critiques on their construct validity.} As well as referencing psychology literature to aid the developments of measurement techniques \cite{yalcin2019evaluating}, we should be familiar with existing evaluations and discussions on their construct validity. Future work is needed to investigate the validity of our current approaches methodologically.

\mysubsection{The Narrow Task Problem: Lessons from our own field.} The interaction between cognitive and emotional processes is significant. Cognitive processes can enable higher empathic accuracy. Perspective-taking is a method of cognitive empathy that enables better empathic understanding, and the appraisal theory of empathy provides a framework specific to situational aspects an observer can consider. These processes relate clearly to reasoning processes explored by NLP tasks. We recommend future work that intersects with the newer area of \textit{abductive commonsense reasoning} \cite{bhagavatula2020abductive}. 

\mysubsection{Refocusing our efforts: Supporting clinical and educational domains aligns better with the motivations of the empathy research area.}
Much of what motivates the empathy research area is a desire to support those in need (i.e., \textit{compassionate} motivations, to employ the new terminology). Our position is that current efforts in empathetic dialogue generation should be reallocated to support the needs of clinical domains, e.g., systems that support communicative skill training for people in care-providing roles and systems for supportive cognitive empathic skills development in-general.\footnote{Ethical perspective is provided in the ethics section.} The need for cognitive empathy is even more pressing for these applications, and therefore NLP research can best support these needs by emphasizing the cognitive aspects in our empathy conceptualizations going forward.

\section{Conclusion}

Language technology is deployed increasingly in interpersonal settings that require an empathetic understanding of the speaker.
Our field's construct represents a significant obstacle to advancing our ability to develop empathetic language technologies. While multiple NLP works have attempted to incorporate empathy into their models and applications, we argue that these works have underspecified empathy---focusing primarily on emotion---and overlooked a major component: cognitive empathy. However, we argue that this gap represents a significant opportunity for developing new models that better reflect cognitive processes and theory of mind while supporting much-needed human-centered applications, such as those in clinical settings.

\section*{Ethical Perspectives}

While our paper is largely an argument for increased depth and specificity in the study of NLP, our call to overcome these obstacles still comes with ethical considerations. Following, we outline two main ethical points.

\cameraready{\mysubsection{Adopt data statements for improving the Definition and Measurement problems and addressing the ethical implications of empathy and NLP.} The research area suffers from low quality and quantity of data, and is no exception from tendency to disproportionately focus on the English language, as there are fewer datasets developed and utilized in other languages (see list in Appendix~\sref{appendix:datasets}). Thus, there is work to be done in data curation. While we are actively encouraging research in praxis-related areas, we shall scrutinize potential approaches for both construct validity and their implications for populations affected by the data. We argue the community should \emph{adopt the practice of creating data statements} \cite{bender2018data}, which will also help with addressing the definition and measurement problems.}

The conversational settings used for empathetic communication often feature highly personal dialog from the target that require special care, e.g., those in the medical domain. Advocating for more work in empathy comes at a potential risk to these targets in keeping their potentially-sensitive data private outside of valid uses. 
The sensitive nature of the data also likely increases the costs and difficulty of developing empathy resources, rendering such efforts rather infeasible for many \cite{hosseini2021takes, welivita-etal-2021-large}.
Furthermore, some datasets cannot be made public to uphold license agreements to protect the rights and privacy of the stakeholders, which is especially the case in counseling domains \cite{althoff-etal-2016-large,perez-rosas-etal-2017-understanding,sharma-etal-2020-computational,zhang-danescu-niculescu-mizil-2020-balancing,zhang2020quantifying}. While these efforts to preserve privacy are rightly stringent, they also create a two-tier system where only researchers who have access to subjects can use the data to participate in such research. Thus, we call for consideration of possible initiatives to enable researchers to utilize datasets crafted for domains beyond publishable social media data. For instance, shared tasks (such as WASSA 2021 and 2022 \cite{tafreshi-etal-2021-wassa,barriere-etal-2022-wassa}) may enable many to experiment on carefully crafted datasets. 
Initiatives may draw inspiration from recent CLPsych shared task models that enabled approved researchers to work on sensitive data under signed data use and ethical practice agreements.\footnote{\url{https://clpsych.org/sharedtask2022/}}

Prior work has, for the most part, assumed that empathy must be prosocial and, therefore, improved empathy models would offer societal benefits. However, emotional and cognitive understanding is not always employed for prosocial purposes. For instance, understanding emotions can be used for abuse, and manipulation \cite{hart1995hare,hodges2007balancing}. Empathy is also capable of motivating hostile acts and fueling hostility toward out-groups~\cite{breithaupt2012three,breithaupt-2018-bad-empathy}; or, if an observer engages with a target with whom they have a bad relationship, empathy for the target's negative experiences may lead to ``malicious gloating''  on the observer's behalf \citep[][p.~259]{bischof1991development}. Thus, the development of new empathetic systems and data could lead to uses that are decidedly antisocial. Consider also the role of empathy in persuasion. In a prosocial context, one study on advice-giving robots showed that students were more likely to be persuaded by the advice of the robot when it used empathetic strategies \cite{langedijk2021more}. 
Emotional appeals are an effective rhetorical tool for persuasion, such as through personal narratives that can increase understanding of other perspectives \cite{wachsmuth-etal-2017-computational,vecchi-etal-2021-towards}. But on that end, emotional understanding can also be used for manipulative appeals to emotion. \citet{huffaker-et-al-2020-eml} introduced a task of detecting emotionally manipulative language, language intended to induce an emotional reaction in the reader (e.g., fear-mongering rhetoric to spread). Their study reviews how adversarial actors have strategically used emotionally manipulative rhetoric in media manipulation. NLP attention on abstract ideas of empathy could be dedicated to such problems. For instance, a new setup could be imagined to anticipate emotional empathy to emotionally manipulative stimuli or focus more on the target's language, which leads to pathogenic empathy in the observers.

\section*{Limitations}

We presented theoretical groundwork on empathy from psychology and neuroscience. We provide distinct descriptions of empathy-related concepts in a way we view practical for aligning research methodologies with more precise conceptualizations for the NLP community. The disambiguation approach we presented is informed by our efforts to review and understand multiple perspectives from psychology and neuroscience thoroughly. We ultimately constructed this approach with references to selected studies from fields that helped us to identify meaningful delineations between the concepts. However, there is no single conceptualization of empathy or related concepts from those areas, so our construct will differ from some. We hope our paper as a whole inspires collective efforts from our community to scrutinize the empathy frameworks that ground the empathy research direction with more informed perspectives. 

This work presented themes about empathy descriptions and evaluations based on a systematic literature review. The search methodology is limited in that it focuses primarily on papers in the ACL Anthology. Our review of related works includes literature outside of ACL venues, and they are consistent with the themes we identify. However, a more extensive study, including works from a broader set of publication venues using standard survey methodology, would provide a more comprehensive outlook on empathetic technology research.

\section*{Acknowledgements}
This work has been supported by the German Federal Ministry of Education and Research (BMBF) as a part of the Junior AI Scientists program under the reference 01-S20060, the Alexander von Humboldt Foundation, and by Hessian.AI. Any opinions, findings, conclusions, or recommendations in this material are those of the authors and do not necessarily reflect the views of the BMBF, Alexander von Humboldt Foundation, or Hessian.AI.

\bibliography{custom,anthology}
\bibliographystyle{acl_natbib}

\appendix

\section{Definition Themes}\label{appendix:definition_themes}

\mysubsection{(5) Thorough, theory-based description of empathy.} Empathy is extensively described with the inclusion of perceptive insights from psychology, neuroscience, or other related fields. There are clear distinctions between empathy and concepts such as emotion recognition and mirroring, sympathy, and compassion. There may be explicit efforts to describe and distinguish between emotional and cognitive empathy and the significance of the distinction.
\textbf{Examples:} \citet{sharma-etal-2020-computational}; \citet{wambsganss-etal-2021-supporting}; \citet{sabour2021cem}.

\mysubsection{(4) Description is succinct yet explicitly theory-grounded and adheres to the theoretical foundation.} Empathy is explicitly defined or described in a way grounded in a complex psychological perspective. As opposed to \textbf{(5)}, longer, more specified descriptions of empathy and its dimensions are not provided. Nevertheless, these works remain consistent with the selected perspective through their analyses and interpretations of their findings. 
\textbf{Example:} \citet{perez-rosas-etal-2017-understanding}.

\mysubsection{(3) Thorough descriptions of other concepts and their relations to empathy are consistent with multi-dimensional theories.} Empathy itself is not explicitly defined or described. It may be unclear whether the idea is rooted in psychology with no specific references. However, behaviors associated with empathy are described that are consistent with multiple aspects of empathy described in the Theoretical Guide \sref{sec:theory}. This category can also include studies that do not specifically describe (or explore) the concept of empathy, but relate empathy to other behaviors (e.g., counselor strategies) in a way that is based on psychology literature. There may also be descriptions of concepts that are not specifically conveyed as empathy, but suggest an independent thought process that arrived at perspectives consistent with multiple aspects reviewed in \sref{sec:theory}.
\textbf{Examples:} \citet{ito-etal-2020-relation}; \citet{zhang-danescu-niculescu-mizil-2020-balancing}; \citet{wu-etal-2021-towards}; \citet{gao-etal-2021-improving-empathetic}.

\mysubsection{(2) Descriptions provided are vague or ambiguous with other concepts.} The descriptions may go briefly beyond the ``understanding the user and responding emotionally" concept by mentioning other aspects (e.g., perspective-taking) in passing or by describing empathetic responses to be those that mimic or mirror the target's response. Other concepts may be mentioned in a way that lacks separable distinction. Some papers reference a psychology definition but leave one or more (usually cognitive) components untouched in their work~\cite{khanpour-etal-2017-identifying}. This theme typically emerges when the study derives a conception based on their own reasoning~\cite{shen-etal-2021-constructing-emotional,zhu-etal-2022-multi}, which could benefit by incorporating terminology and distinctions we provide in \sref{sec:theory}. 

\mysubsection{(1) No empathy description but abstract conceptualization may be inferred through task or system description.} No definition nor description of empathy or empathetic behaviors is explicitly stated, however some abstract conceptualization of empathy can be inferred through the description of the task or requirements of an empathetic system. This includes describing the empathetic response generation task as ``to understand the user emotion and respond appropriately" \citet{lin-etal-2019-moel,rashkin-etal-2019-towards}. Elaboration on the process of understanding the user's emotion in an empathetic way or responding appropriately in an empathetic way are not provided.

\mysubsection{(0) No empathy description, and the conceptualization is not inferrable through other information.} No definition of empathy is provided, despite the apparent relevance of empathy to the study. Such works may perform a human evaluation of empathy, and an emotional perspective on empathy may be inferred through a description of the evaluation if provided~\cite{phy-etal-2020-deconstruct}. Generally, the conceptualization cannot be inferred. This label does not include works that reflect abstract conceptualization in their task description as in Theme 1. This category includes cases where empathy may be ambiguous with concepts such as ``politeness" and ``courtesy''~\cite{firdaus-etal-2020-incorporating}.

\section{Evaluation Themes}\label{appendix:evaluation_categories}

Here we categorize papers by how they approached evaluating, labeled, or rated empathy in their research results or when constructing datasets.

\mysubsection{Multi-item: cognitive \& emotional (8).} These studies report manual tasks for labeling or rating multiple items (e.g., behaviors, strategies) of empathy and include cognitive and emotional aspects in these items. This category includes studies that developed a new scheme for labeling or rating response intents and behaviors, or labeled counselor behaviors based on schemes designed for a counseling style. In a setup particularly unique among the NLP literature, one recruited participants to complete a multi-item self-report scale for trait empathy (Davis’ Interpersonal Reactivity Index (IRI) \cite{davis1980-multidim-empathy,davis1983measuring}) and then analyzed linguistics behaviors with respect to those scales \cite{litvak-etal-2016-social}. 
\textbf{All papers:} \citet{litvak-etal-2016-social};
\citet{perez-rosas-etal-2017-understanding};
\citet{sharma-etal-2020-computational};
\citet{welivita-pu-2020-taxonomy};
\citet{ito-etal-2020-relation};
\citet{zhang-danescu-niculescu-mizil-2020-balancing};
\citet{wambsganss-etal-2021-supporting};
\citet{svikhnushina-etal-2022-taxonomy}.

\mysubsection{Single label/rating: cognitive \& emotional empathy (3).} These studies report manual tasks for labeling or rating empathy as a single item or score based on multiple items or aspects representing both cognitive and emotional empathy. Two are based on theoretical and practical psychology conceptualizations (Appraisal Theory of Empathy \cite{zhou-jurgens-2020-condolence} and MISC \cite{wu-etal-2021-towards}). The other is based on a description of empathy that includes intents or acts by the observer to help regulate the target’s emotions \cite{sanguinetti-etal-2020-annotating}. 
\textbf{All papers:} \citet{zhou-jurgens-2020-condolence};
\citet{sanguinetti-etal-2020-annotating};
\citet{wu-etal-2021-towards}.

\mysubsection{Single label/rating: emotional empathy (19).}\label{appendix:emotional_empathy} These studies report manual tasks for labeling or rating empathy as a single item or score based on a description of empathy (reported to be provided to the annotators) representing only emotional aspects of empathy. Task setups include comparisons between two items (which item is more empathetic based on the provided description of empathy), binary labels (empathetic or not), and ratings on different Likert scales. The typical descriptions of empathy provided are defined by whether or the degree to which the observer item shows, demonstrates, or expresses an ability to infer or an understanding/awareness of the target’s emotions or feelings. Often these descriptions include that the observer should respond in a way that is appropriate, emotionally or otherwise, without guidelines on appropriate vs. inappropriate empathetic responses. Some descriptions refer to emotion sharing, such as saying the observer should manifest, share, or experience the target’s emotions \cite{zhu-etal-2022-multi}. Some studies appear to use sympathy and empathy interchangeably \cite{rashkin-etal-2019-towards, lin-etal-2019-moel}. We determined this category to be the best fit for \citet{buechel-etal-2018-modeling}'s study, in which the single empathy ratings are based on a multi-item questionnaire focused on emotions.
\textbf{All papers:} \citet{alam2016can};
\citet{alam-etal-2016-interlocutors};
\citet{alam2018annotating};
\citet{buechel-etal-2018-modeling};
\citet{rashkin-etal-2019-towards};
\citet{lin-etal-2019-moel};
\citet{smith-etal-2020-put};
\citet{majumder-etal-2020-mime};
\citet{naous-etal-2020-empathy};
\citet{phy-etal-2020-deconstruct};
\citet{li-etal-2020-empdg};
\citet{zeng-etal-2021-affective};
\citet{wu-etal-2021-transferable};
\citet{shen-etal-2021-constructing-emotional};
\citet{xie-pu-2021-empathetic};
\citet{sabour2021cem};
\citet{zheng-etal-2021-comae};
\citet{naous-etal-2021-empathetic};
\citet{zhu-etal-2022-multi}.

\mysubsection{Single label/rating: no specification (8).} These studies refer to or report a manual task performed to label or rate empathy without a clear description of empathy upon which the task was based. Task setups include asking annotators to label items as ``empathic responses'' and to compare two items of observer text (empathetic response generation output) for which is ``more empathetic.'' Four of these studies ask annotators to rate empathy on Likert scales (3-point, 4-point, 5-point, and 7-point). A few studies indicated they recruited annotators with linguistics or psychology backgrounds \cite{tu-etal-2022-misc}. One study describes a procedure for familiarizing the annotators with the concept based on selected papers from psychology and unifying their understanding with the help of psychologists \cite{khanpour-etal-2017-identifying}. 
\textbf{All papers:} \citet{abdul2017recognizing};
\citet{khanpour-etal-2017-identifying};
\citet{kim-etal-2021-perspective};
\citet{gao-etal-2021-improving-empathetic};
\citet{chen-etal-2021-nice-neural};
\citet{ishii-etal-2021-erica};
\citet{jang-etal-2021-bpm};
\citet{tu-etal-2022-misc}.

\mysubsection{Heuristic empathy labels or ratings (4).} These studies include heuristic methods to gather or label empathetic data automatically. One study created an ``empathy lexicon'' based on associations with single-item empathy ratings on a prior dataset \cite{sedoc-etal-2020-learning}. Another built a dataset by training a model on data labeled with scheme of empathetic response intents \cite{welivita-etal-2021-large}. One study selected conversations from two subreddits (r/happy and r/offmychest) and manually labeled samples from them and a control group (r/CausalConversations) as empathetic or non-empathetic (1 or 0), and compared the averages of the selected subreddits to the control to demonstrate that they were more empathetic on average \cite{zhong-etal-2020-towards}. Another considered an ``empathetic language style'' to be the language style of ``more empathetic'' of two Myers-Briggs Type Indicator (MBTI) personality types \cite{vanderlyn-etal-2021-seemed}. We labeled \citet{rashkin-etal-2019-towards}'s study based on their human evaluation of the empathetic response generation model. However, we also consider their data curation, in which crowd-workers have conversations grounded on specific emotions, to be a heuristic-based approach.
\textbf{All papers:} \citet{sedoc-etal-2020-learning};
\citet{zhong-etal-2020-towards};
\citet{welivita-etal-2021-large};
\citet{vanderlyn-etal-2021-seemed}.

\mysubsection{Role labeling (2).} Instead of labeling, rating, or associating behaviors with empathy, these studies label target and observer roles (seeker vs. provider \cite{hosseini-caragea-2021-distilling-knowledge} and seeking empathy vs. providing empathy \cite{hosseini2021takes}).
\textbf{All papers:} \citet{hosseini2021takes};
\citet{hosseini-caragea-2021-distilling-knowledge}.

\mysubsection{Only automatic or no manual evaluation (4).} This category includes studies that report empathetic systems or results without reporting manual evaluation \cite{siddique-etal-2017-zara,zhou-etal-2020-design,firdaus-etal-2020-incorporating} or that evaluate empathy automatically only \cite{tsai-etal-2021-style}. 
\textbf{All papers:} \citet{siddique-etal-2017-zara};
\citet{zhou-etal-2020-design};
\citet{firdaus-etal-2020-incorporating};
\citet{tsai-etal-2021-style}.

\mysubsection{Buechel/WASSA (15).} All papers: \citet{fornaciari-etal-2021-milanlp}; \citet{tafreshi-etal-2021-wassa}; \citet{butala-etal-2021-team}; \citet{vettigli-sorgente-2021-empna}; \citet{mundra-etal-2021-wassa}; \citet{kulkarni-etal-2021-pvg}; \citet{guda-etal-2021-empathbert}; \citet{qian-etal-2022-surrey}; \citet{chen-etal-2022-iucl}; \citet{del-arco-etal-2022-empathy}; \citet{vasava-etal-2022-transformer}; \citet{ghosh-etal-2022-team}; \citet{lahnala-etal-2022-caisa}; \citet{barriere-etal-2022-wassa}.

\mysubsection{Not categorized (38).} Not empathy task or does not evaluate. All papers:
\citet{suarez-etal-2012-building}; \citet{castellano2013towards}; \citet{bhargava-etal-2013-demonstration}; \citet{denis-etal-2014-synalp}; \citet{fung-etal-2016-zara}; \citet{hastie2016remember}; \citet{fung-etal-2016-zara-supergirl}; \citet{addawood-etal-2017-telling}; \citet{iserman-ireland-2017-dictionary}; \citet{hazarika-etal-2018-icon}; \citet{hazarika-etal-2018-conversational}; \citet{guerini-etal-2018-methodology}; \citet{zhou-wang-2018-mojitalk}; \citet{mahajan-shaikh-2019-emoji}; \citet{demasi-etal-2019-towards}; \citet{demszky-etal-2020-goemotions}; \citet{shen-feng-2020-cdl}; \citet{shuster-etal-2020-dialogue}; \citet{inoue-etal-2020-attentive};  \citet{wang-etal-2020-contextualized}; \citet{roller-etal-2021-recipes}; \citet{hu-etal-2021-dialoguecrn}; \citet{varshney-etal-2021-modelling}; \citet{yoo-etal-2021-empathy}; \citet{inoue-etal-2021-multi}; \citet{guo-choi-2021-enhancing}; \citet{lu-etal-2021-retrieve-discriminate}; \citet{hu-etal-2021-mmgcn}; \citet{li-etal-2022-continuing}; \citet{maheshwari-varma-2022-ensemble}; \citet{falk-lapesa-2022-reports}; \citet{ide-kawahara-2022-building}; \citet{bhandari-goyal-2022-bitsa}; \citet{stephan2015empathy}; \citet{langedijk2021more}; \citet{pruksachatkun2019moments}; \citet{sabour2021cem}.

\section{Emotion recognition, contagion, and mimicry}\label{appendix:contagion_mimicry}
Two underlying processes of emotional empathy, emotion recognition, and contagion, are closely related \cite{shamay2011neural}. \textit{ Emotional contagion} (``catching feelings'') refers to when an observer's brain activates similarly to a target's (e.g., perception of pain), where the observer lacks awareness of the origin of their emotion \cite{decety2006human,ickes2011everyday}. Research on the human mechanism of imitating an affective state~\cite{goldman-1993-ethics,carr-et-al-2003} relates to some response generation approaches in NLP~\cite{majumder-etal-2020-mime}. However, mimicking emotions without experiencing contagion is a separate concept referred to as ``mimpathy''~\cite{ickes2011everyday}.

\section{Non-English Datasets} Arabic \citealp{naous-etal-2021-empathetic}, Italian \citealp{alam2018annotating,sanguinetti-etal-2020-annotating}, Chinese \citealp{sun-etal-2021-psyqa}, Japanese \citealp{ito-etal-2020-relation,sanguinetti-etal-2020-annotating}, German \citealp{wambsganss-etal-2021-supporting}.

\end{document}